# Machine Learning for Air Transport Planning and Management

Graham Wild[1]
*University of New South Wales, Canberra, ACT, 2612, Australia*

Glenn Baxter[2] and Pannarat Srisaeng[3]
*Suan Dusit University, Bangkok, 10300, Thailand*

Steven Richardson[4]
*Edith Cowan University, Joondalup, WA, 6027, Australia*

In this work we compare the performance of several machine learning algorithms applied to the problem of modelling air transport demand. Forecasting in the air transport industry is an essential part of planning and managing because of the economic and financial aspects of the industry. The traditional approach used in airline operations as specified by the International Civil Aviation Organization is the use of a multiple linear regression (MLR) model, utilizing cost variables and economic factors. Here, the performance of models utilizing an artificial neural network (ANN), an adaptive neuro-fuzzy inference system (ANFIS), a genetic algorithm, a support vector machine, and a regression tree are compared to MLR. The ANN and ANFIS had the best performance in terms of the lowest mean squared error.

## I. Introduction

The world's airlines play a critical role in facilitating global commerce, tourism, and merchandise trade. In the global airline passenger industry, there are four distinct business models: full-service network carrier (FSNC), low-cost carrier (LCC), regional carrier, and holiday/charter airline [1]. Regardless of the chosen business model, forecasting is regarded as one of the most critical management functions for an airline [2]. Passenger forecasts are used by airlines for a variety of important purposes. Airlines forecast their passenger demand so they can optimally plan the supply of services that will be necessary to satisfy that demand [2]. Passenger forecasts are also used for financial purposes by investors as investment efficiency is significantly influenced by the accuracy and adequacy of the estimation performed by the airline [3]. Passenger air traffic forecasts are also one of the critical inputs into an airline's fleet plan, and their route network development. The forecasts are also used by the airline when preparing the firm's annual operating plan [2, 4]. Furthermore, the analysis and forecasting passenger air travel demand can be of considerable assistance to an airline in reducing its risk through an objective evaluation of the demand side of its passenger business [4]. There are also a variety of other purposes for an airline's passenger forecasts. Passenger forecasts provide a basis for the evaluation of staffing and facilities requirements throughout an airlines network. In addition, passenger forecasts are used to develop the airline's marketing strategies as well as its promotional programs. Finally, passenger forecasts can be used by airlines to identify the travelling public's future requirements, desires, and propensity to travel and transport goods by air [5].

While currently, the airline industry is recovering from the impacts of the COVID-19 pandemic, historically the scale and growth of the global commercial air transportation has been significant. In 2019, the average passenger flew 1,936 km, and there were 4.486 billion passengers; the product of these gives the total Revenue Passenger Kilometres

---

[1] Senior Lecturer, School of Engineering and Information Technology, AIAA Member.
[2] Associate Research Fellow, School of Tourism and Hospitality Management.
[3] Assistant Professor, School of Tourism and Hospitality Management.
[4] Senior Lecturer, School of Science.

(RPKs) as 8.686 trillion [6]. The trend in RPKs since the end of World War 2 (1945), in addition to Freight Tonne Kilometers (FTKs), is shown in Fig. 1. An FTK is the cargo equivalent of RTKs, given as the product of the number of tonnes and the distance flown. While traffic (both cargo and passengers) has increased exponentially, the economic reality of the airline industry does not match this. Fig. 2 shows the past twenty years of revenue and profit, which shows an average profit margin of 1.03% for the airline industry, which for companies in the S&P 500 have average 8.4%. These narrow profit margins in a capital-intensive industry (an Airbus A320 was listed at $101 million in 2018, with the A350 averaging $320 million [7]), means that forecasting is a critical part of planning and management. It should be noted that clearly the years including the COVID-19 pandemic have been excluded, although including these only makes the sensitivity of the airline industry more apparent.

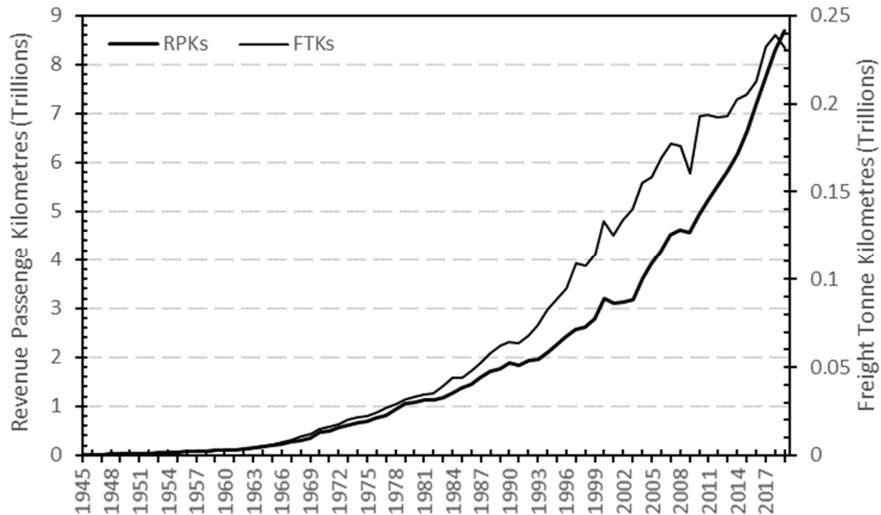

**Fig. 1 Growth of the global aviation industry since 1945, both passenger RPKs and cargo FTKs.**

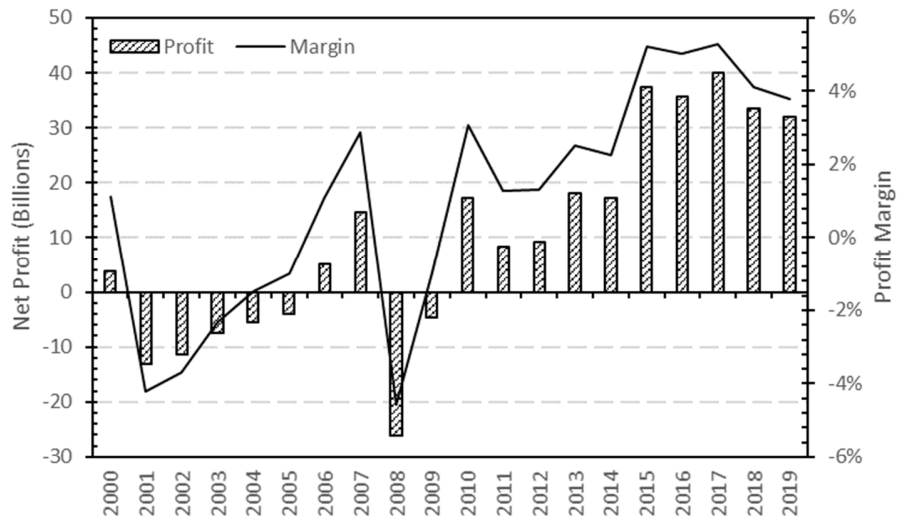

**Fig. 2 Total annual profits reported by airline industry (in billions of USD) and the corresponding margin (percentage), from 2000 to 2019.**

The objective of this study is to examine the evolution in the approaches that are used to forecast airline passenger demand and to identify the approach (method) that provides the greatest statistical accuracy. To achieve this objective a real-world case study based on Australia's low-cost carrier passenger market is examined in which the traditional multiple linear regression method is compared with five machine learning based methods, specifically, artificial neural networks, genetic algorithms, adaptive neuro-fuzzy inference systems, support vector machines, and regression trees.

The remainder of the paper is structured as follows. Section 2 presents an overview of the studies that have employed multiple linear regression (MLR) to forecast airline passenger demand. This is followed in Section 3 by a review of the various reported studies that have applied machine learning based approaches to predict airline passenger demand. The real-world case study is presented in Section 4. The key findings of the study are presented in Section 5.

## II. The Traditional Airline Passenger Forecasting Approach

In the global air transport industry, many key stakeholders and government regulatory bodies base their passenger forecast on the International Civil Aviation Organization's (ICAO) Manual on Air Traffic Forecasting [8], originally developed by ICAO in 1985. Historically, multiple linear regression (MLR) models have typically been used to forecast air traffic demand. In 2000, Ba-Fail et al. [4] developed an MLR model for analysing the determinates of domestic air travel demand in the Kingdom of Saudi Arabia. Not long after in 2001, Abed et al. [9] developed an econometric model for predicting international air travel demand in the Kingdom of Saudi Arabia. Also, in 2001, ICAO developed econometric models to forecast global air passenger growth to 2010 [10]. In 2010, Aderamo [11] proposed and tested MLR-based models for forecasting the demand for domestic air transport in Nigeria. Kopsch [12] in 2012 developed a model for forecasting the demand for domestic air travel in Sweden. In 2013, Sivrikaya and Tunç [13], proposed and tested models for forecasting domestic air travel demand in Turkey. In more recent research, Adenirano and Stephens [14] proposed and empirically tested models for forecasting Nigeria's international airline passenger demand. Adeniran and Kanyio [15] in a further study used Ordinary Least Square (OLS) regression to forecast the ten years (2018 to 2028) demand for international air passenger travel in Nigeria. Kim [16] proposed MLR models for forecasting international airline demand in Korea.

## III. The Evolution in Artificial Intelligence Passenger Forecasting Approaches

In contrast to the traditional MLR approach, there has been a growing body of studies that have applied artificial intelligence methods to predict air travel demand. In 2002, Alekseev and Seixus [17] developed an artificial neural network (ANN)-based forecasting model to predict the annual Brazilian air transport passenger demand. This was followed by a similar study by the same authors in 2009 that proposed and applied an ANN to predict the annual Brazilian air transport passenger demand [18]. Blinova [3] developed an ANN to forecast the development of Russia's air transport network. In further ANN-related work, Chen et al. [19] employed a back-propagation artificial neural network (BPN) to improve the forecasting accuracy of air passenger and air cargo demand from Japan to Taiwan. Al-Rukaibi and Al-Mutairi [20] modelled the factors influencing air travel demand in Kuwait using the traditional linear regression models and an artificial neural network (ANN) approach. The authors concluded that the ANN was the model of choice due to the better goodness of fit measures over the regression model results. Srisaeng et al. [21] developed and tested an ANN for predicting Australia's domestic passenger and revenue passenger kilometres performed (RPKs). Following this Srisaeng et al. [22] estimated Australia's low-cost carrier demand using a traditional MLR model and an ANN. The authors found that the ANN had a higher predictive capability than the MLR models. Kolidakis and Botzoris [23] developed an ANN to forecast Sydney Airport airline passenger demand, using data covering the period January 2005 until August 2018. In a further study conducted in 2018, Koç and Arslan [24] developed and tested an ANN for forecasting domestic air transport in Turkey between 2007 and 2015. Dingari et al. [25] used an ANN to forecast Air India's domestic passengers from 2012 to 2018.

There have also been several reported studies that have proposed and empirically tested genetic algorithms for predicting aviation-related demand. Sineglazov et al. [26] proposed a new algorithm, which was based on the group method of data handling, and an artificial neural network. The authors' algorithm was tested on real data and showed better results than the ANNs. Srisaeng et al. [27] developed two genetic algorithm-based models - GAPAXDE (enplaned passengers) and GARPKSDE (revenue passenger kilometres performed model) – for predicting Australia's domestic air travel demand. In related work, the authors further developed two GA-based models - GAPAXDE Model (enplaned passengers) and GARPKSDE Model (revenue passenger kilometres performed). The modelling results showed that both the linear GAPAXDE and GARPKSDE models were more accurate, reliable, and offered a slightly greater predictive capability as compared to the quadratic models [28]. Mohie-Eldin et al. [29] used an ANN and a genetic algorithm approach to forecast the air passenger demand in Egypt (International and Domestic passengers). The period examined in the study was from 1970 to 2013. Alarfaj and Al Ghowinem [30] in 2018 applied different forecasting models: genetic algorithm, artificial neural network, and classical linear regression to forecast Saudi Arabia's domestic LCC passenger demand.

Srisaeng et al. extended their use of AI-methods for predicting air travel demand in two further studies which have estimated Australia's low-cost carrier (LCC) [31] and regional air travel demand [32] using adaptive neuro-fuzzy inference systems (ANFIS), respectively. The results of both studies showed that the ANFIS provided a very high predictive capability.

### IV. Case Study of Australia's Low Cost Carrier Market

#### A. The Case

A Low Cost Carrier (LCC) is an airline that eliminates non-essential services in order to reduce the cost of the air fare [33]. What this means in practice is that the ticket price only includes the cost of the air fare, and everything else is treated as an optional extra. This includes assigned seating, checked baggage, meals, drinks, entertainment, and other amenities [34]. There is also typically only a single cabin class (no premium seating), maximizing the capacity for passengers. The reduction in cost not only applies to the passenger, it applies to the fleet and operational support. Fleets are simple utilizing where possible a single common aircraft type, helping to reduce cost for flight and cabin crews (everyone is familiar and certified for the entire fleet), maintenance (engineers and spares are only needed for a single type), and ground handling (tugs, tow bars, stairs, belt loaders etc can all be interchanged and are uniform) [35]. Sales are typically handled using online booking systems, and at airports, self-service check in and bag drop reduce the need for check in staff.

The operation of a LCC also has unique features. Services tend to be from secondary airports [36], which offer lower fees, and reduced traffic which can help reduce delays. Delays are a significant issue in an airline network where the goal is to have every aircraft operational at the same time, for as long as possible, to maximize revenue. That is, time on the ground is minimized, with turnaround times as short as possible [37]. The operational network is also different to legacy carriers that focus on business travel. Clearly a legacy network is intended to connect business centers, while an LCC network connects population centers and holiday destinations, in specific direct links, not via a hub.

While Compass Airlines can be identified as the first LCC in Australia in the early 1990's [38], the business model took off in the 2000's with Impulse Airlines and Virgin Blue. While Impulse was acquired by the legacy carrier Qantas, Virgin Blue succeeded, as the second legacy carrier, Ansett Australia collapsed, and it was part of Virgin Group, so they had the requisite financial backing to weather the financially turbulent industry illustrated in Fig. 2. In 2004, Qantas launched Jetstar Airways, Australia's second successful LCC. The final major Australian LCC was Tiger Airways, which commenced operation in 2007, and in 2012 was acquired by Virgin Australia. Virgin Australia was competing with Qantas and saw Tiger Airways as an option to compete with Jetstar. Tiger ceased operating in 2020 due to the impacts of the COVID-19 pandemic.

#### B. Modelling

Previous work has looked at modeling the LCC traffic demand in Australia [22, 28, 31]. As the goal of this work is not to specifically model and forecast a specific key aviation parameter with a specific machine learning tool; rather, it is to compare different machine learning tools for a given aviation parameter. The machine learning tools compared in this work include artificial neural networks (ANNs), genetic algorithms (GAs), adaptive neuro-fuzzy inference systems (ANFIS), support vector machines (SVM), and regression trees (RT). These are all compared to each other, and the baseline multiple linear regression model. For each, the same data points (hence proportions) were used for training, testing, and validation.

#### C. Variables and Data

Previous work includes details of the variables selected and the sources of the data [22, 28, 31]. For completeness, the variables included in all models were:
- Australia's real best discount economy air fare (a poxy for yield),
- Australia's Gross Domestic Product (GDP) per capita,
- Australia's unemployment numbers,
- Australia's interest rate,
- World jet fuel prices,
- Australia's tourist accommodation capacity,
- Dichotomous variables for the effect of 9/11,
- Dichotomous variables for the change in Virgin's business model,
- Dichotomous variables for the Sydney 2000 Olympic Games
- Dichotomous variables for the Melbourne 2006 Commonwealth Games

**D. Results**

Fig. 3 to Fig. 8 show the results for each of the six models in terms of the predicted RPKs relative to the actual RPKs for all 42 quarters of data. The order of the graphs is, MLR, GA, ANN, ANFIS, SVM, and RT. Each figure includes the equation (which would ideally give β = 1, α = 0), which gives an indication of the accuracy of the modelling tool, while the included $R^2$ value is an indication of the precision of the modelling tool. To provide a side-by-side comparison, Table 1 includes the relevant root mean squared error (RMSE) terms, in order of best to worse. The ANFIS model resulted in the lowest RMSE, while the RT has the highest. The significant error associated with the RT is illustrated in Fig 8.

**Table 1 RMSE terms for each modelling tool in billions of RPKs and as a percentage of the 2019 total**

|            | ANFIS | ANN  | GA   | MLR  | SVM  | RT   |
|------------|-------|------|------|------|------|------|
| RMSE(RPKs) | 136   | 173  | 267  | 301  | 317  | 350  |
| % of 2019  | 1.57  | 1.99 | 3.07 | 3.47 | 3.65 | 4.03 |

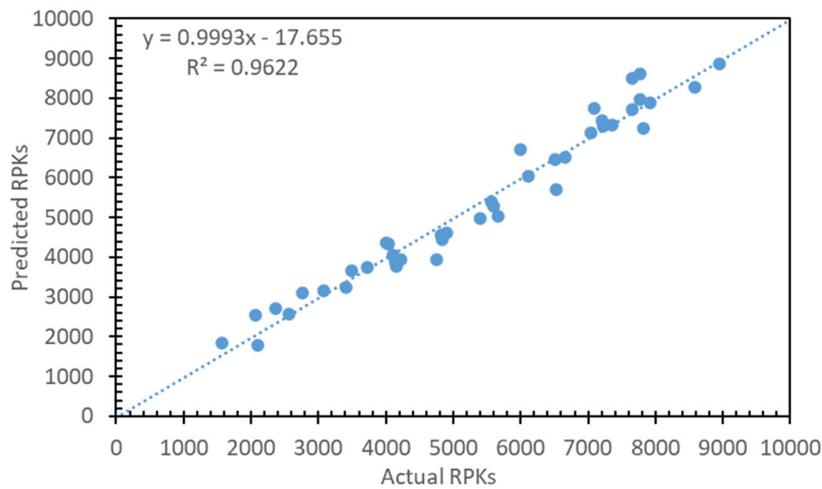

Fig. 3  MLR results for all 42quarters of data.

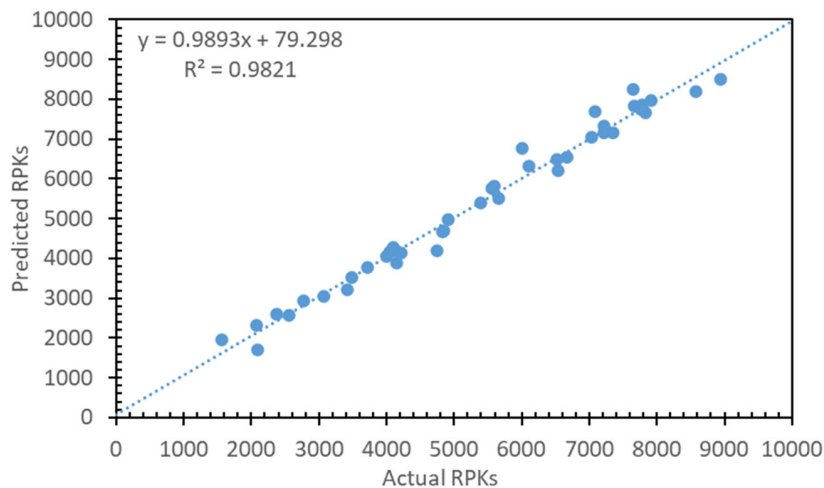

Fig. 4  GA results for all 42quarters of data.

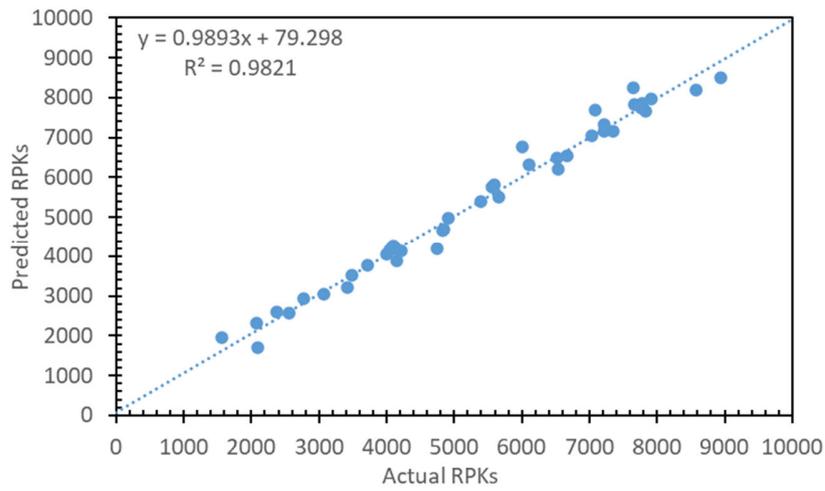

Fig. 5  ANN results for all 42quarters of data.

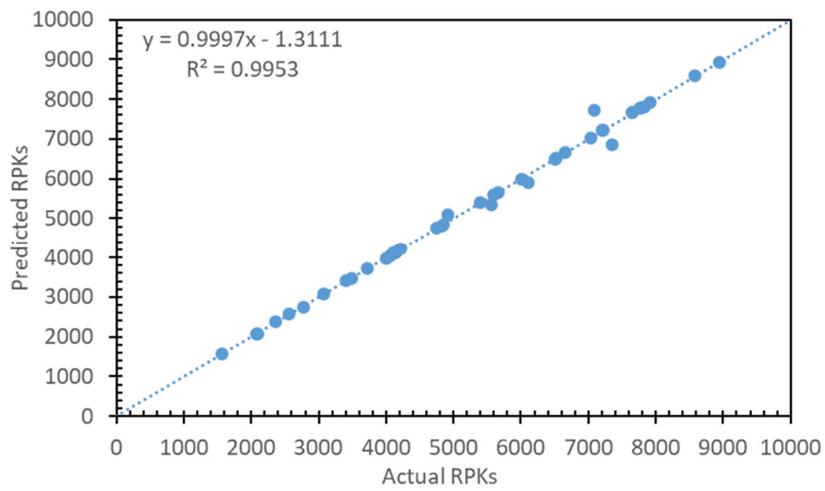

Fig. 6  ANFIS results for all 42quarters of data.

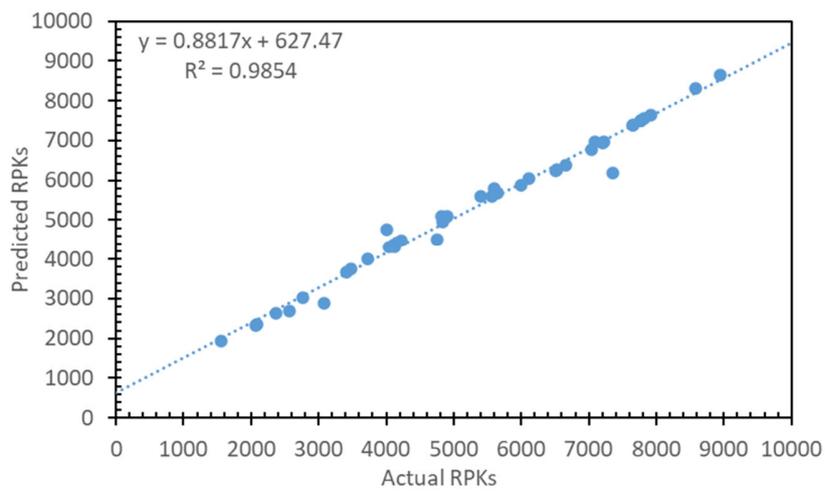

Fig. 7  SVM results for all 42quarters of data.

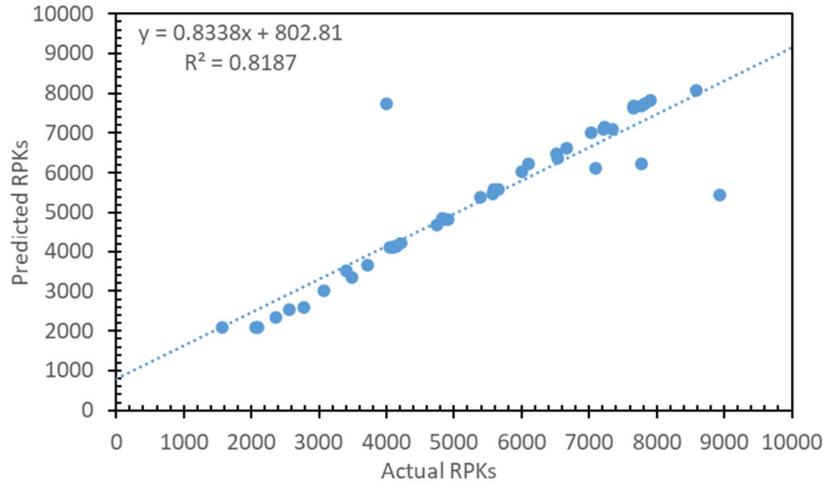

**Fig. 8 RT results for all 42quarters of data.**

To fully assess the models requires the use of ANOVA, comparing all 6 sets of MSE terms, to determine if they are statistically significantly different. As shown in Table 2, the ANOVA test is significant, with $p$ less than 0.05 (corresponding to a 95% confidence level). As a result of the statistical significance, post hoc single-sided two-sample $t$-tests were required to determine which models are statistically significantly different to each other. These results are shown in Table 3. Both the ANN and ANFIS models result in MSE terms that are statistically significantly better than all the other models. Relative to each other, the resultant errors mean that the ANFIS model is not statistically significantly better than the ANN model. The largest error is produced by the RT model means that one cannot conclude that it is statistically significantly worse than all of the other models; however, it is noted that all of the associated $p$-values are less than 0.10 (90% confident), with values of 0.056 and 0.059 relative to the ANFIS and ANN models, respectively.

**Table 2 ANOVA Testing results for MSE terms**

|  | SS | df | MS | F | p |
|---|---|---|---|---|---|
| Between | $1.6 \times 10^{13}$ | 5 | $3.1 \times 10^{12}$ | 2.287 | 0.047 |
| Within | $3.3 \times 10^{14}$ | 246 | $1.4 \times 10^{12}$ |  |  |
| Total | $3.5 \times 10^{14}$ | 251 |  |  |  |

**Table 3 Post hoc single-sided two-sample t-tests for MSE terms, given as p-values**

|  | GA | ANN | ANFIS | SVM | RT |
|---|---|---|---|---|---|
| MLR | 0.210 | 0.002* | <0.001* | 0.396 | 0.076 |
| GA | - | 0.039* | 0.009* | 0.219 | 0.070 |
| ANN | - | - | 0.266 | 0.025* | 0.059 |
| ANFIS | - | - | - | 0.010* | 0.056 |
| SVM | - | - | - | - | 0.080 |

## V. Conclusions

The forecasting of passenger demand is viewed as one of the most critical airline management functions. Using a real-world case study approach, this research has examined various machine learning based tools that have been applied to model Australia's low-cost carrier passenger demand. Based on the resultant root mean squared error (RMSE) to evaluate the models, the study found that the ANFIS approach had the lowest RMSE, whilst the Regression Tree approach had the highest RMSE. The results for the ANFIS were only marginally better than those for an ANN, and the results were not statistically significantly different. The study also concludes that the application of artificial intelligence-based methods can be successfully applied in the airline industry to aid in operational planning and managing.